\documentclass[conference]{IEEEtran}

\usepackage[utf8]{inputenc}
\usepackage{todonotes}
\usepackage{amsmath}
\usepackage{amssymb} 
\usepackage{textcomp} 
\usepackage{gensymb} 

\DeclareMathOperator*{\argmin}{arg\,min}
\newcommand{\norm}[1]{\left\lVert#1\right\rVert}

\begin{document}

\title{3D Face Tracking and Texture Fusion in the Wild}

\author{
\IEEEauthorblockN{Patrik Huber, William Christmas, Josef Kittler}
\IEEEauthorblockA{Centre for Vision, Speech and Signal Processing\\
University of Surrey\\
Guildford, GU2 7XH, United Kingdom\\
Contact: http://www.patrikhuber.ch}
\and
\IEEEauthorblockN{Philipp Kopp, Matthias Rätsch}
\IEEEauthorblockA{Image Understanding and Interactive Robotics\\
Reutlingen University\\
D-72762 Reutlingen, Germany}
}

\maketitle

\begin{abstract}
We present a fully automatic approach to real-time 3D face reconstruction from monocular in-the-wild videos. 
With the use of a cascaded-regressor based face tracking and a 3D Morphable Face Model shape fitting, we obtain a semi-dense 3D face shape. We further use the texture information from multiple frames to build a holistic 3D face representation from the video frames.
Our system is able to capture facial expressions and does not require any person-specific training.
We demonstrate the robustness of our approach on the challenging \emph{300 Videos in the Wild} (300-VW) dataset. Our real-time fitting framework is available as an open source library at 
\url{http://4dface.org}.

\end{abstract}

\section{Introduction}
This paper addresses the problem of reconstructing a 3D face from monocular in the wild videos. While the problem has been studied in the past, existing algorithms rely either on RGB-D data or have not demonstrated their robustness on realistic in-the-wild videos.

From the algorithms working on monocular video sequences, the algorithm of Garrido et al.~\cite{mpi_monocular_siggraphasia_2013} requires manual, subject-specific training and labelling, and their algorithm has only been evaluated on a limited set of videos under rather controlled conditions, with frontal poses and on videos in HD quality. Ichim et al.~\cite{ichim_dynamic_2015} also require subject specific training and manual labelling by an experienced labeler, taking 1 to 7 minutes per subject, which is a tedious process, and their resulting model is person-specific.
Jeni et al.~\cite{jeni_dense_2015} use rendered 3D meshes to train their algorithm, which do not contain the variations that occur in 2D in-the-wild images, for example, the meshes have to be rendered on random backgrounds. Furthermore, they only evaluate by cross-validation on the same 3D data their algorithm has been trained on.
Cao et al.~\cite{cao_3d_shape_2013,cao_dde_2014} reconstruct only shape, without using texture, and do not evaluate on in-the-wild videos. Cao et al.~\cite{DBLP:journals/tog/CaoBZB15} don't require user specific training, but present only results in controlled, frontal and high image resolution and require CUDA to achieve real-time performance.

In contrast to these approaches, we present an approach that requires no subject specific training and evaluate it on a challenging 2D in-the-wild video data set. We are the first to carry out such an evaluation of a 3D face reconstruction algorithm on in-the-wild data with challenging pose and light variations as well as limited resolution and show the robustness of our algorithm. While many of these works focus on face reenactment, we focus on a high-quality texture representation of the subject in front of the camera.

In addition to subject-specific manual training being a tedious step, creating a personalised face model \emph{offline} is not possible where the subject can not be seen beforehand, e.g. for face recognition in the wild, customer tracking for behaviour analysis or various HCI scenarios. Our approach runs in near real-time on a CPU.

This paper presents the following contributions. By combining cascaded regression with 3D Morphable Face Model fitting, we obtain real-time face tracking and semi-dense 3D shape estimate from low-quality consumer 2D webcam videos. We present an approach to fuse the face texture from multiple video frames to yield a holistic textured face model. We demonstrate the applicability of our method to in-the-wild videos on the newly released 300-VW video database that includes challenging scenarios like speeches and TV shows. Furthermore, we present preliminary results of a median-based super-resolution approach that can be applied when the whole video is available in advance.
Finally, our method is available as open-source software on GitHub.

\begin{figure}[t]
\begin{center}
    \includegraphics[width=0.95\linewidth]{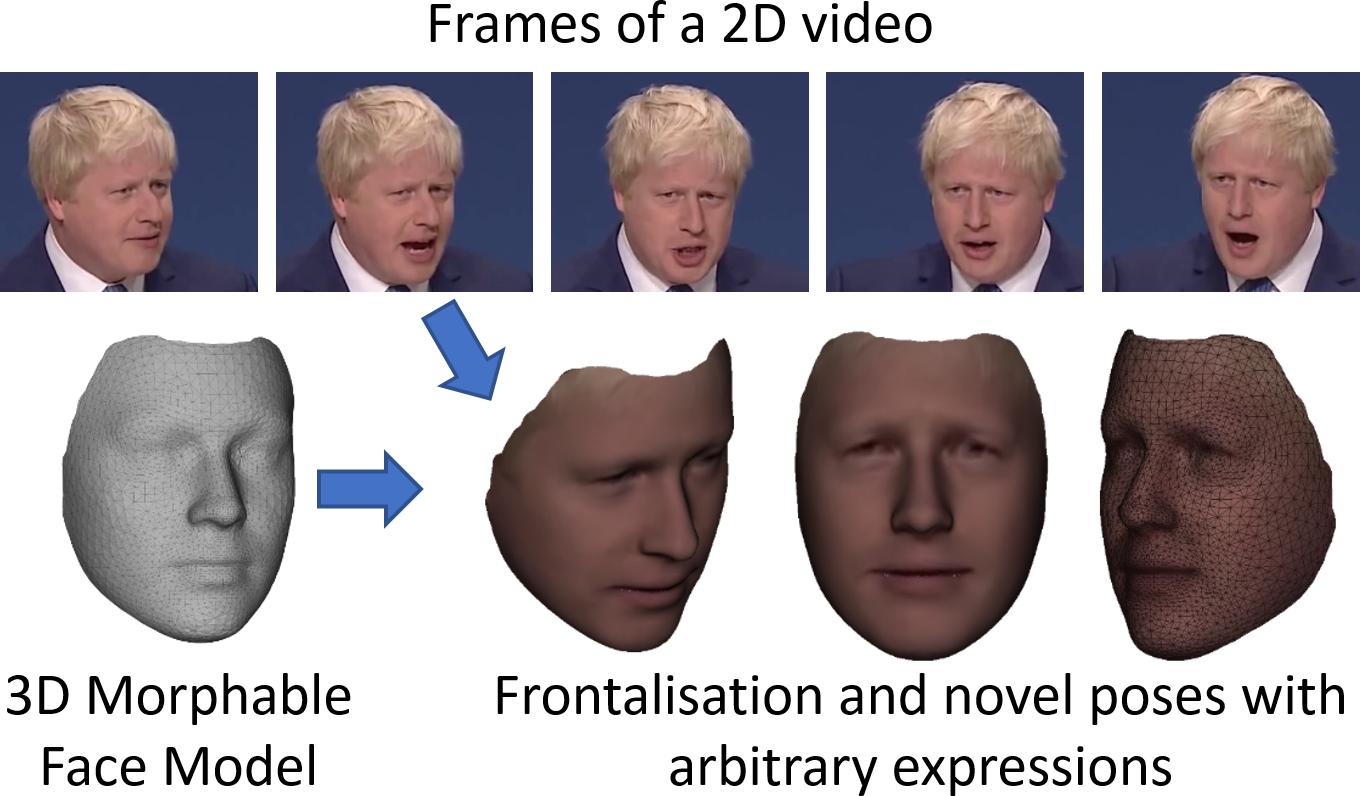} 
\end{center}
\caption{Real-time 3D face reconstruction from a monocular in-the-wild video stream. Our method uses a 3D Morphable Face Model as face prior and fuses the information from multiple frames to create a holistic 3D face reconstruction, without requiring subject-specific training.}
\label{fig:demo}
\end{figure}

\section{Method} \label{sec:method}

In general, reconstructing a 3D face from 2D data is an ill-conditioned problem. To make this task feasible, our approach incorporates a 3D Morphable Face Model (3DMM) as prior knowledge about faces.

In this section, we first briefly introduce the 3D Morphable Face Model we use. We then present our 3D face reconstruction approach and the texture fusion.

\subsection{3D Morphable Face Model}

A 3D Morphable Model (3DMM) is based on three dimensional meshes of faces that have been registered to a reference mesh, i.e. are in dense correspondence. A face is represented by a vector $S \in \mathbb{R}^{3N}$, containing the x, y and z components of the shape, and a vector $T \in \mathbb{R}^{3N}$, containing the per-vertex RGB colour information. $N$ is the number of mesh vertices. The 3DMM consists of two PCA models, one for the shape and one for the colour information, of which we only use the shape model in this paper.
Each PCA model
\begin{equation}
\text{M}:=(\mathbf{\bar{v}}, \boldsymbol{\sigma}, \mathbf{V})
\end{equation}
consists of the mean of the model, $\mathbf{\bar{v}} \in \mathbb{R}^{3N}$, a set of principal components $\mathbf{V} = [\mathbf{v}_1,\hdots,\mathbf{v}_{M}] \in \mathbb{R}^{3N\times M}$, and the standard deviations $\boldsymbol{\sigma} \in \mathbb{R}^{M}$. $ M $ is the number of principal components present in the model.
Novel faces can be generated by calculating
\begin{equation}
S = \mathbf{\bar{v}} + \sum_{i}^{M}\alpha_i\sigma_i\mathbf{v}_i
\label{eq:generate_face}
\end{equation}
for the shape, where $\boldsymbol\alpha \in \mathbb{R}^M$ are a set of 3D face instance coordinates in the shape PCA space.

\subsection{Face Tracking} \label{sec:face_tracking}

To track the face in each frame of a video, we use a cascaded-regression based approach, similar to Feng et al.~\cite{feng_RCRC_2015}, to regress a set of sparse facial landmark points. The goal is to find a regressor:
\begin{equation}
\label{single_regressor}
R: \mathbf{f}(\mathbf{I},\boldsymbol\theta) \rightarrow \delta \boldsymbol\theta,
\end{equation}
where $\mathbf{f}(\mathbf{I},\theta)$ is a vector of extracted image features from the input image, given the current 
model parameters, and $\delta \boldsymbol\theta$ is the 
predicted model parameter update. This mapping is learned from a training dataset using a series of linear regressors

\begin{equation}
\label{linear_regression}
R_n: \delta \theta = \mathbf{A}_n\mathbf{f}(\mathbf{I},\theta) + \mathbf{b}_n,
\end{equation}
where $\mathbf{A}_n$ is the projection matrix and $\mathbf{b}_n$ is the 
offset (bias) of the $n$th regressor.
and $\mathbf{f}(\mathbf{I},\theta)$ extracts HOG features from the image.

When run on a video stream, the regression is initialise at the location from the previous frame but with the model's mean landmarks, which acts as a regularisation.

\subsection{3D Model Fitting} \label{sec:model_fitting}

In subsequent steps, the 3D Morphable Model is fitted to the subject in a frame. This section describes our camera model, the PCA shape fitting, and subsequent refinement using contour landmarks and facial expressions. 

\noindent{\textbf{Camera model}\hspace{5pt}}
With the 2D landmark locations and their known correspondences in the 3D Morphable Model, we estimate the pose of the camera. We assume an affine camera model and implement the \textit{Gold Standard Algorithm} of Hartley \& Zisserman~\cite{hartley_multiple_2004}, which finds a least squares approximation of a camera matrix given a number of 2D - 3D point pairs.

First, the detected 2D landmark points $ x_i \in \mathbb{R}^3 $ and the corresponding 3D model points $ X_i \in \mathbb{R}^4 $ (both represented in homogeneous coordinates) are normalised by similarity transforms that translate the centroid of the image and model points to the origin and scale them so that the Root-Mean-Square distance from their origin is $\sqrt{2}$ for the landmark and $\sqrt{3}$ for the model points respectively: $ \tilde{x}_i = \mathbf{T}x_i $ with $ \mathbf{T} \in \mathbb{R}^{3\times3}$, and $ \tilde{X}_i = \mathbf{U}X_i $ with $ \mathbf{U} \in \mathbb{R}^{4\times4} $. Using $ \geq 4 $ landmark points, we then compute a normalised camera matrix $ \tilde{\mathbf{C}} \in \mathbb{R}^{3\times4} $ using the \textit{Gold Standard Algorithm} \cite{hartley_multiple_2004} and obtain the final camera matrix after denormalising: $ \mathbf{C} = \mathbf{T}^{-1}\tilde{\mathbf{C}}\mathbf{U} $.

\medskip
\noindent{\textbf{Shape fitting}\hspace{5pt}}
Given the estimated camera, the 3D shape model is fitted to the sparse set of 2D landmarks to produce an identity-specific semi-dense 3D shape.  We find the most likely vector of PCA shape coefficients $\boldsymbol{\alpha}$ by minimising the following cost function:
\begin{equation}
	\mathbb{E} = \sum_{i=1}^{3N} \frac{(y_{m2D,i} - y_{i})^{2}}{2\sigma^2_{2D}} + \lVert \boldsymbol\alpha \rVert_2^2 \,,
	\label{eq:shape_fitting}
\end{equation}
where $N$ is the number of landmarks, $y$ are detected or labelled 2D landmarks in homogeneous coordinates, $\sigma^2_{2D}$ is an optional variance for these landmark points, and $y_{m2D}$ is the projection of the 3D Morphable Model shape to 2D using the estimated camera matrix. More specifically, $y_{m2D,i} = \mathbf{P}_i\cdot(\mathbf{\hat{V}}_h \alpha + \mathbf{\bar{v}})$, where $\mathbf{P}_i$ is the $i$-th row of $\mathbf{P}$ and $\mathbf{P}$ is a matrix that has copies of the camera matrix $\mathbf{C}$ on its diagonal, and $\mathbf{\hat{V}}_h$ is a modified PCA basis matrix that consists of a sub-selection of the rows that correspond to the landmark points that the shape is fitted to. Additionally, a row of zeros is inserted after every third row to accommodate for homogeneous coordinates, and the basis vectors are multiplied with the square root of their respective eigenvalue.
The cost function in~(\ref{eq:shape_fitting}) can be brought into a standard linear least squares formulation. For details of the algorithm, we refer the reader to~\cite{aldrian2013inverse} and \cite{huber_visapp_multiresolution_2016}.

\subsection{Expression Fitting}

To model expressions, we use a set of expression blendshapes $ \mathbf{B} $ that have been computed from 3D expression scans. A linear combination of these blendshapes is added to the PCA model, so a shape is represented as:

\begin{equation}
S = \mathbf{\bar{v}} + \sum_{i}^{M}\alpha_i\sigma_i\mathbf{v}_i + \sum_{j}^{L}\psi_j\mathbf{B}_j ,
\label{eq:generate_face_expressions}
\end{equation}
where $ \mathbf{B}_j $ is the $j$-th column of $\mathbf{B}$.

To solve for the blendshape coefficients, we use a non-negative least squares solver that minimises the distance between the current estimated model projection and the 2D landmarks. Because we solve for $ \boldsymbol\alpha $ and $ \boldsymbol\psi $ at the same time, we run the PCA shape and the expression fitting alternating until they reach stable values - usually they converge within ten iterations.

\begin{figure}[ht]
\begin{center}
    \includegraphics[width=0.30\linewidth]{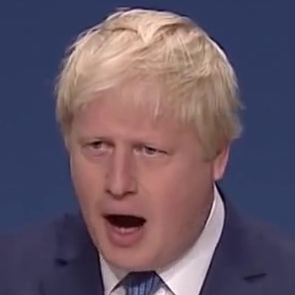}
    \includegraphics[width=0.30\linewidth]{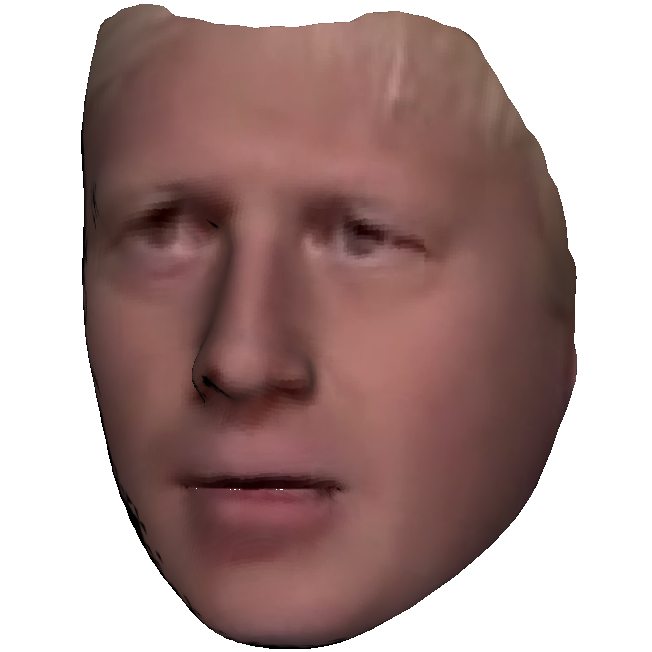}
    \includegraphics[width=0.30\linewidth]{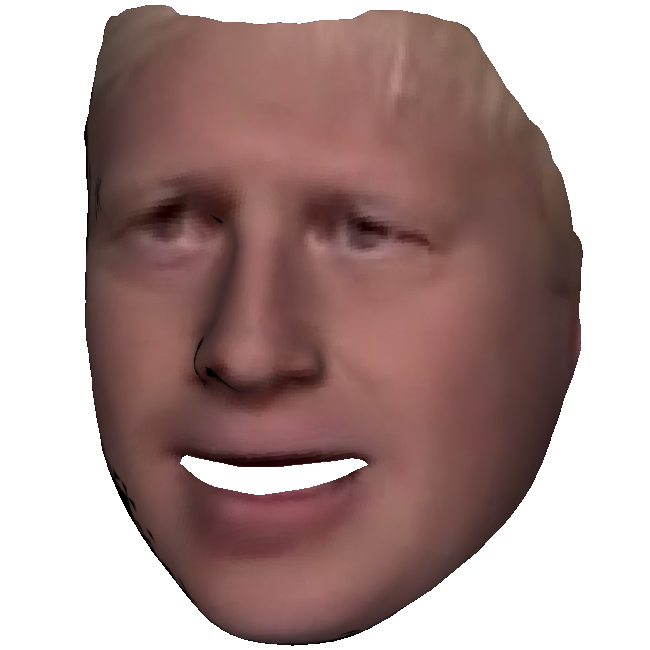}
\end{center}
\caption{Frame with strong expression and expression-neutralised image. \emph{(left)}: Input frame. \emph{(middle)}: Expression-neutralised 3D model. \emph{(right)}: Face with artificially added smile expression.}
\label{fig:expression_neutralisation}
\end{figure}

The result is identity specific shape coefficients $ \boldsymbol\alpha $ and expression blendshapes $ \boldsymbol\psi $. Besides allowing to model expressions present in the subject in front of the camera, it can be used to remove a facial expression from a subject, or to re-render it with a different expression. Figure~\ref{fig:expression_neutralisation} shows a frame with a strong expression, the expression-neutralised face, and a re-rendering with a synthesised expression.

\subsection{Contour Refinement}

In general, the outer face contours present in the 2D image do not correspond to unique contours on the 3D model. At the same time, these contours are important for an accurate face reconstruction, as they define the boundary region of the face. This problem has had limited attention in the research community, but for example Bas et al.~\cite{DBLP:journals/corr/BasSBW16} recently provided an excellent overview describing the problem in more detail.

To deal with this problem of contour correspondences, we introduce a simple contour fitting that fits the front-facing face contour given semi-fixed 2D-3D correspondences. We assume that the front-facing contour (that is, the half of the contour closer to the camera, for example the right face contour when a subject looks to left) corresponds to the outline of the model.
We thus define a set of vertices $ V $ along the outline of the 3D face model, and then, given an initial fit, 
search for the closest vertex in that list for each detected 2D contour point.

Given a 2D contour landmark $ y $, the optimal corresponding 3D vertex $ \hat{v} $ is chosen as:

\begin{equation}
\hat{v} = \argmin_{v\in V} \norm{Pv-y}^2,
\label{eq:contour_correspondence}
\end{equation}
where $ P $ is the currently estimated projection matrix from 3D to 2D.

Using a whole set of potential 3D contour vertices makes the method robust against varying roll and pitch angles, as well as against vertical inaccuracies of the contour from the landmark regressor.
Once these optimal contour correspondences are found, they are used as additional corresponding points in the algorithm described in the previous sections.

\subsection{Texture Reconstruction} \label{sec:texture_reconstruction}

Once an accurate model-fit is obtained, we remap the image texture from a frame to an isomap that puts each pixel 
into a globally registered representation.
The isomap is a texture map, created by projecting the 3D model triangles to 2D while preserves the geodesic distance between vertices. The mapping is computed only once, so the isomaps of each frames are in dense correspondence with each other.

\begin{figure}[ht]
\begin{center}
    \includegraphics[width=0.30\linewidth]{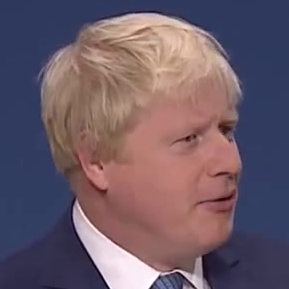}
    \quad
    \includegraphics[width=0.30\linewidth]{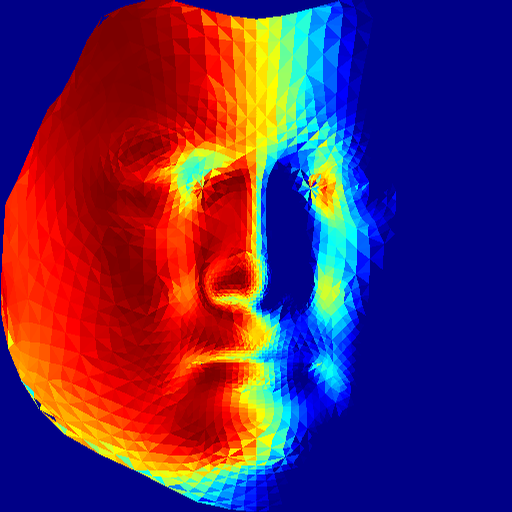}
\end{center}
\caption{View visibility information (including regions of self-occlusions) from the 3D face model. \emph{(left)}: Input frame. \emph{(right)}: red = 0\degree (facing the camera), blue = 90\degree or facing away. JET colourmap.}
\label{fig:view_weighting}
\end{figure}

Inspired by \cite{van_rootseler_using_2012}, we compute a weighting $ \omega $ for each point in the isomap that is given by the angle of the camera viewing direction $ \vec{d} $ and the normal of the 3D mesh's triangle of the current point $ \vec{n} $: $ \omega = \langle \vec{a}, \vec{n} \rangle $. Thus, vertices that are facing away from the camera receive a lower weighting than vertices directly facing the camera, and self-occluded regions are discarded.
In contrast to \cite{van_rootseler_using_2012}, our approach does not depend on the colour model or a colour or light model fitting. 
Figure~\ref{fig:view_weighting} shows an example image and the resulting weighting for each pixel.

To reconstruct the texture value at each pixel location, we calculate a weighted average of all frames up to the current one, each pixel weighed by its triangle's computed $ \omega $ of a particular frame. This average can be computed very efficiently, i.e. by adding the values of the current frame to the previous average and normalising accordingly, without having to recompute the values for all previous frames.
While 
more complex fusion techniques could be applied, 
our method is particularly suited for real-time application and in that it allows the computation of an incremental texture model on a video stream, without having knowledge of the whole video in advance.

\section{Experiments} \label{sec:experiments}

\begin{figure*}[t]
\centering
\includegraphics[width=0.08\linewidth,clip=true,trim=0 0 0 0]{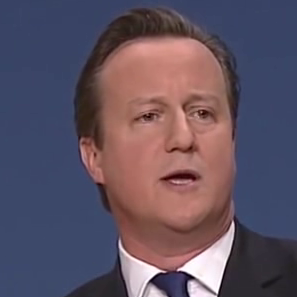} 
\includegraphics[width=0.08\linewidth,clip=true,trim=0 0 0 0]{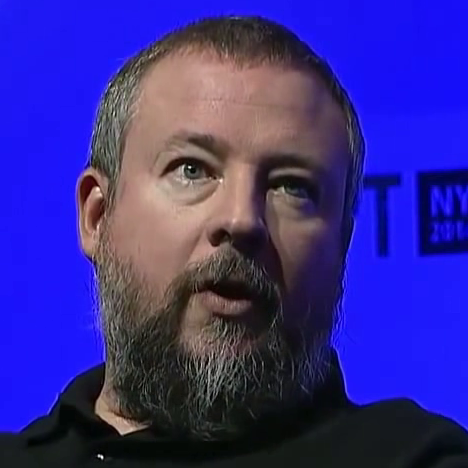} 
\includegraphics[width=0.08\linewidth,clip=true,trim=0 0 0 0]{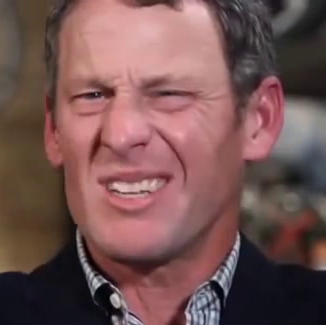} 
\includegraphics[width=0.08\linewidth,clip=true,trim=0 0 0 0]{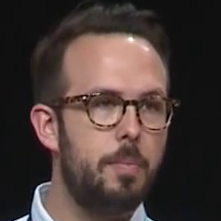} 
\includegraphics[width=0.08\linewidth,clip=true,trim=0 0 0 0]{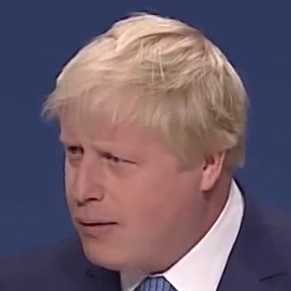} 
\includegraphics[width=0.08\linewidth,clip=true,trim=0 0 0 0]{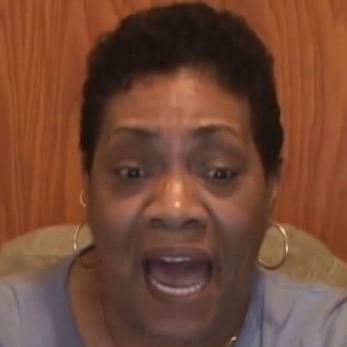} 
\includegraphics[width=0.08\linewidth,clip=true,trim=0 0 0 0]{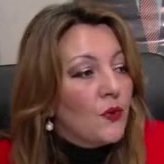} 
\includegraphics[width=0.08\linewidth,clip=true,trim=0 0 0 0]{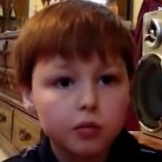} 
\includegraphics[width=0.08\linewidth,clip=true,trim=0 0 0 0]{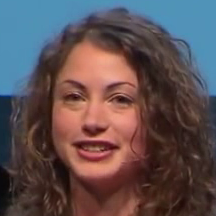} 
\includegraphics[width=0.08\linewidth,clip=true,trim=0 0 0 0]{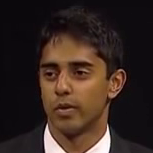} 
\\
\vspace{0.3em}
\includegraphics[width=0.08\linewidth,clip=true,trim=0 0 0 0]{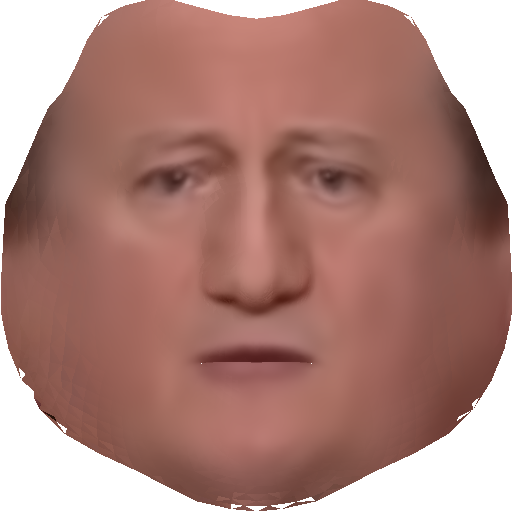} 
\includegraphics[width=0.08\linewidth,clip=true,trim=0 0 0 0]{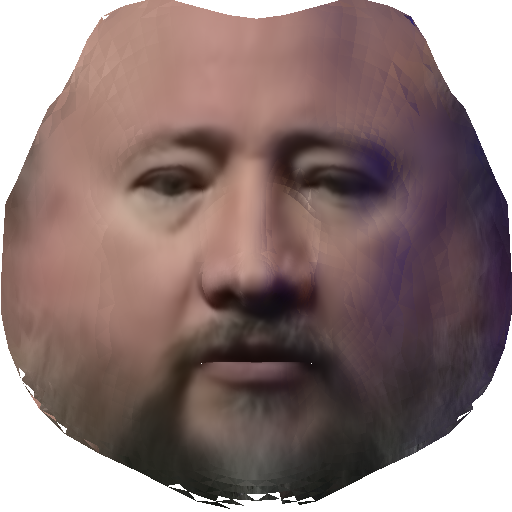} 
\includegraphics[width=0.08\linewidth,clip=true,trim=0 0 0 0]{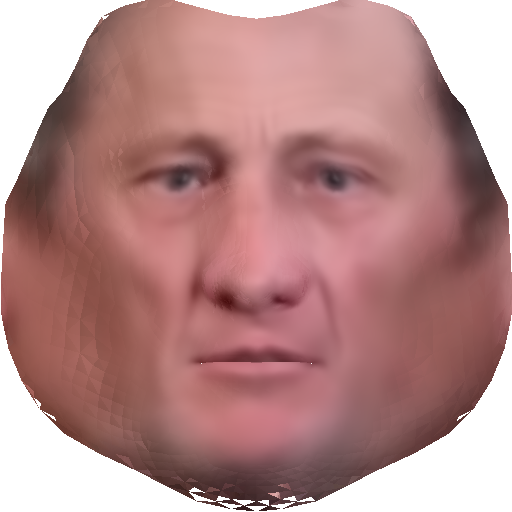} 
\includegraphics[width=0.08\linewidth,clip=true,trim=0 0 0 0]{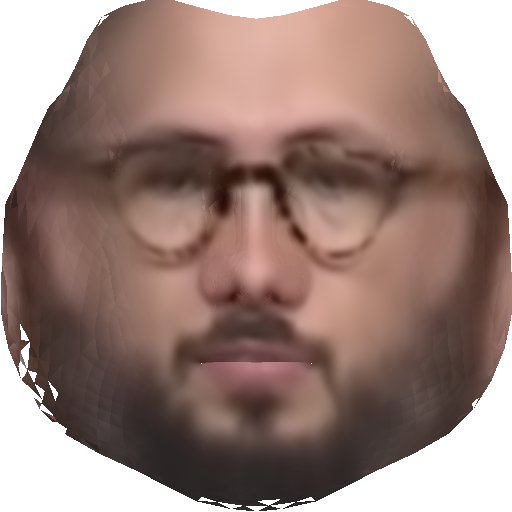} 
\includegraphics[width=0.08\linewidth,clip=true,trim=0 0 0 0]{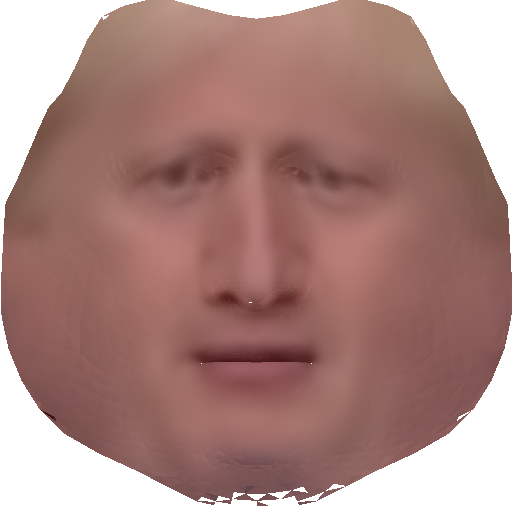} 
\includegraphics[width=0.08\linewidth,clip=true,trim=0 0 0 0]{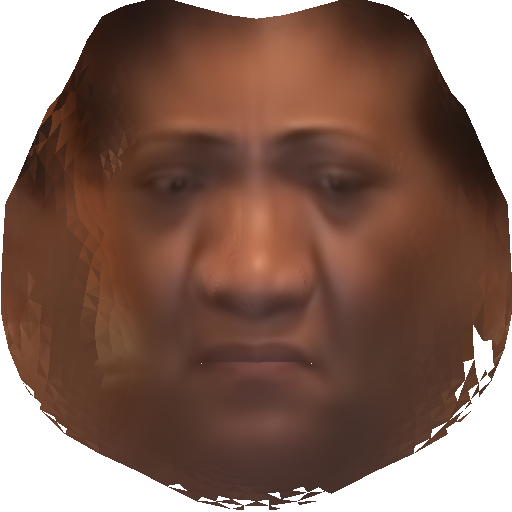} 
\includegraphics[width=0.08\linewidth,clip=true,trim=0 0 0 0]{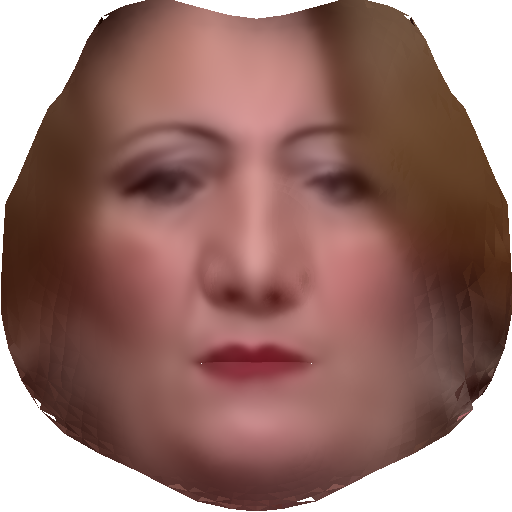} 
\includegraphics[width=0.08\linewidth,clip=true,trim=0 0 0 0]{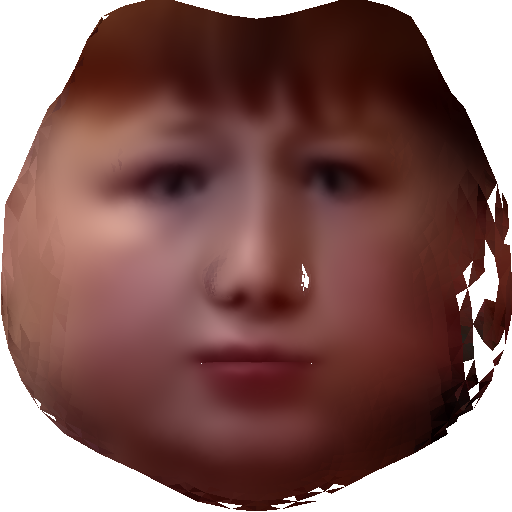} 
\includegraphics[width=0.08\linewidth,clip=true,trim=0 0 0 0]{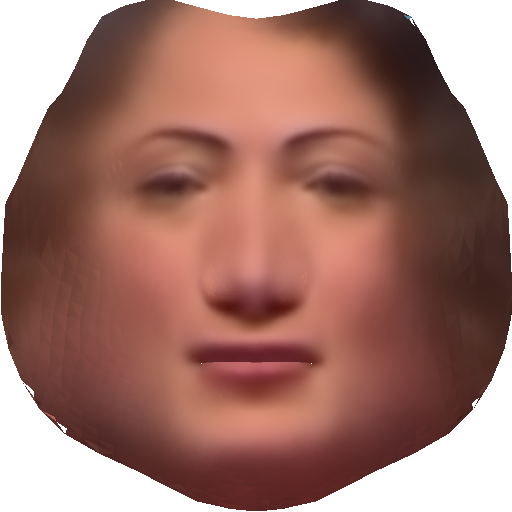} 
\includegraphics[width=0.08\linewidth,clip=true,trim=0 0 0 0]{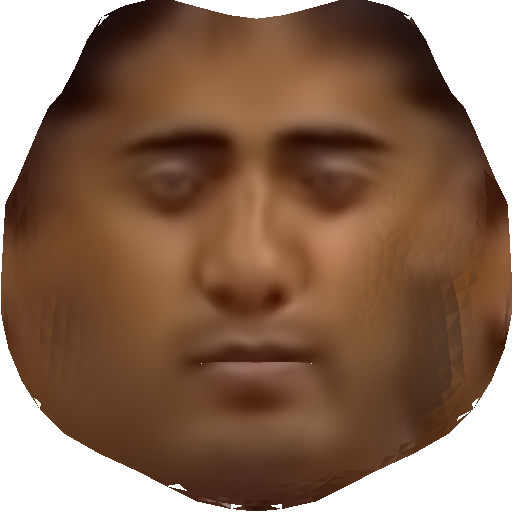} 
\\
\vspace{0.3em}
\includegraphics[width=0.08\linewidth,clip=true,trim=0 0 0 0]{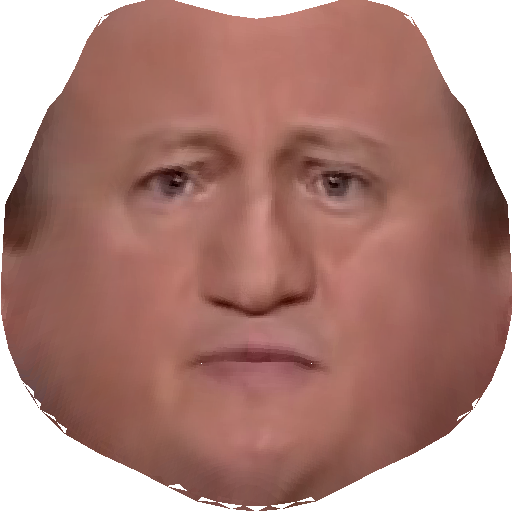} 
\includegraphics[width=0.08\linewidth,clip=true,trim=0 0 0 0]{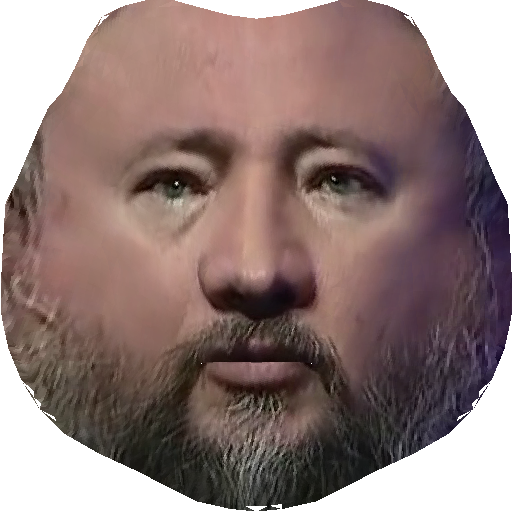} 
\includegraphics[width=0.08\linewidth,clip=true,trim=0 0 0 0]{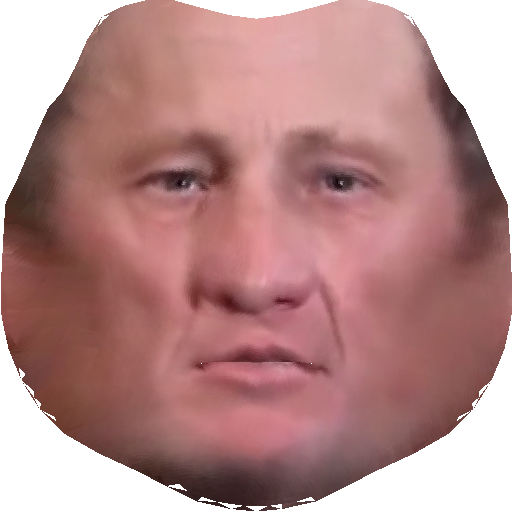} 
\includegraphics[width=0.08\linewidth,clip=true,trim=0 0 0 0]{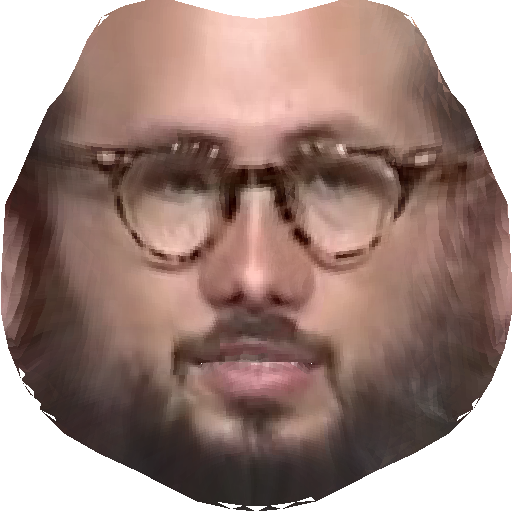} 
\includegraphics[width=0.08\linewidth,clip=true,trim=0 0 0 0]{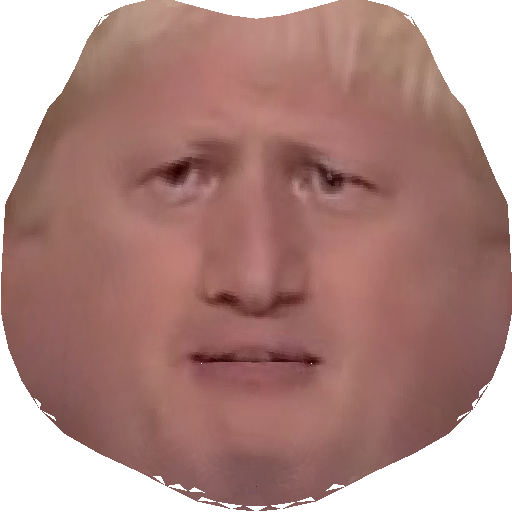} 
\includegraphics[width=0.08\linewidth,clip=true,trim=0 0 0 0]{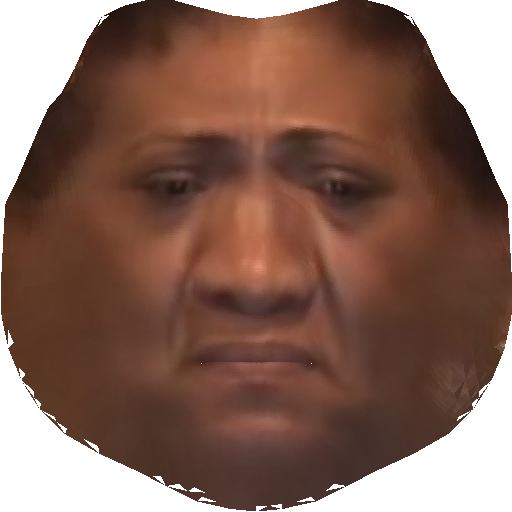} 
\includegraphics[width=0.08\linewidth,clip=true,trim=0 0 0 0]{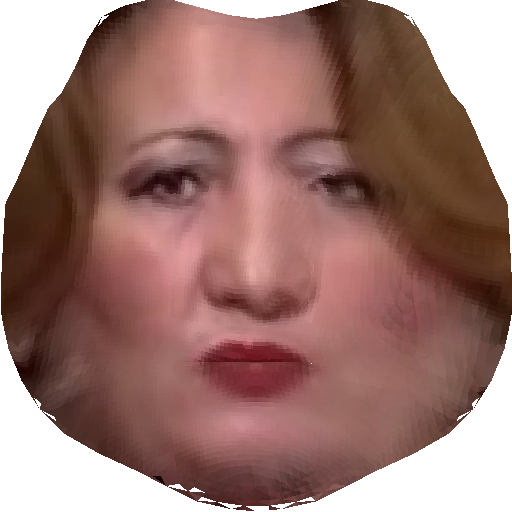} 
\includegraphics[width=0.08\linewidth,clip=true,trim=0 0 0 0]{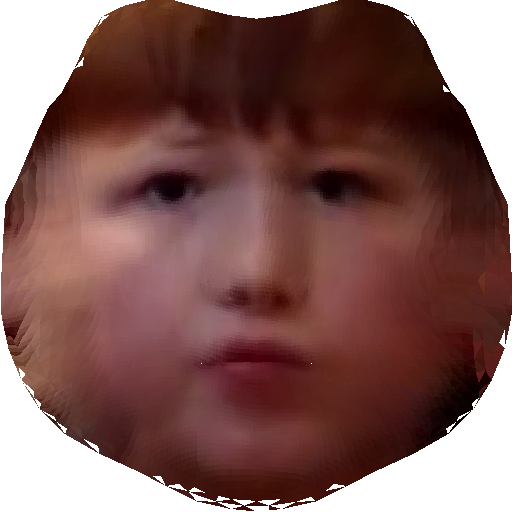} 
\includegraphics[width=0.08\linewidth,clip=true,trim=0 0 0 0]{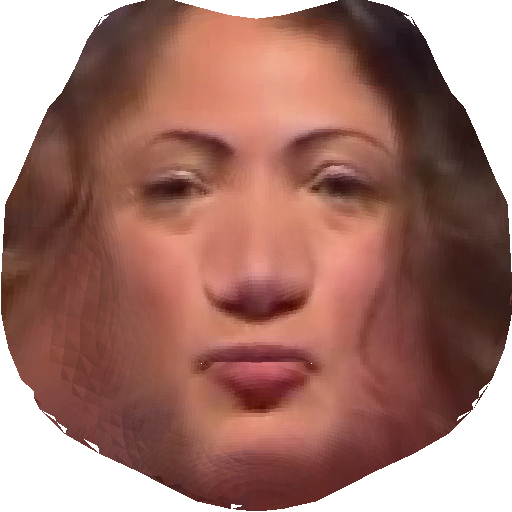} 
\includegraphics[width=0.08\linewidth,clip=true,trim=0 0 0 0]{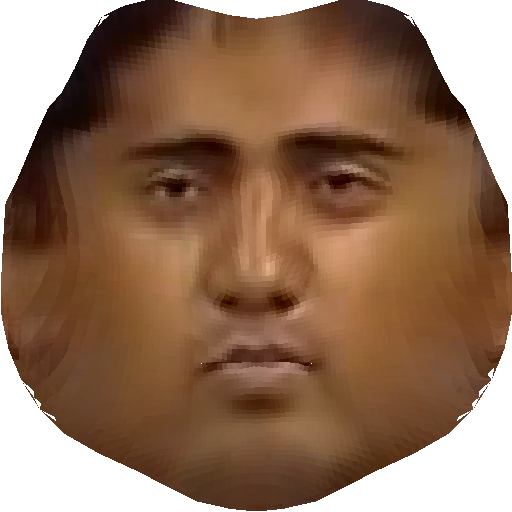} 
\\
\vspace{0.3em}
\includegraphics[width=0.08\linewidth,clip=true,trim=0 0 0 0]{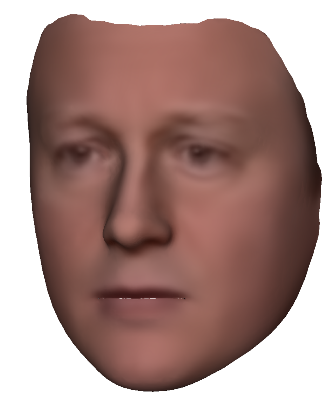} 
\includegraphics[width=0.08\linewidth,clip=true,trim=0 0 0 0]{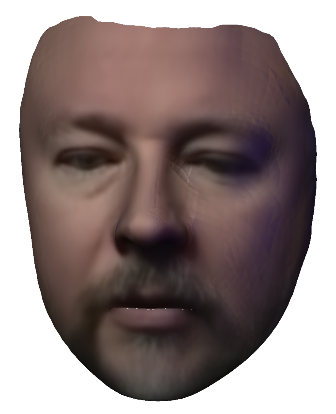} 
\includegraphics[width=0.08\linewidth,clip=true,trim=0 0 0 0]{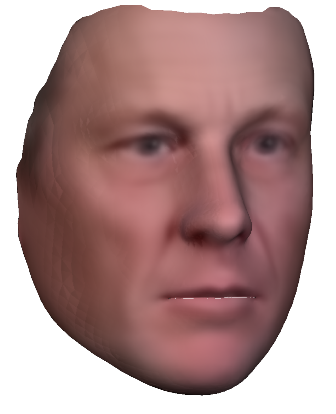} 
\includegraphics[width=0.08\linewidth,clip=true,trim=0 0 0 0]{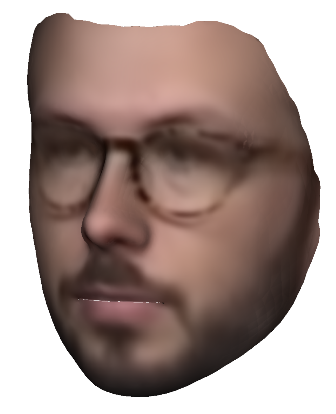} 
\includegraphics[width=0.08\linewidth,clip=true,trim=0 0 0 0]{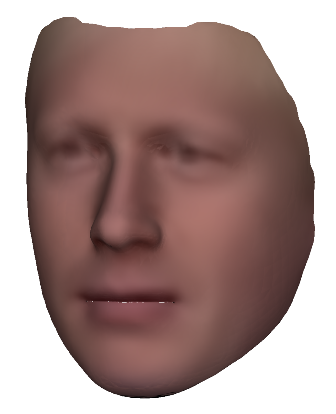} 
\includegraphics[width=0.08\linewidth,clip=true,trim=0 0 0 0]{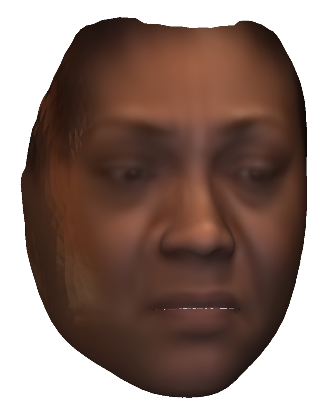} 
\includegraphics[width=0.08\linewidth,clip=true,trim=0 0 0 0]{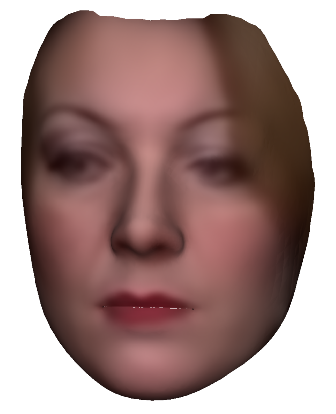} 
\includegraphics[width=0.08\linewidth,clip=true,trim=0 0 0 0]{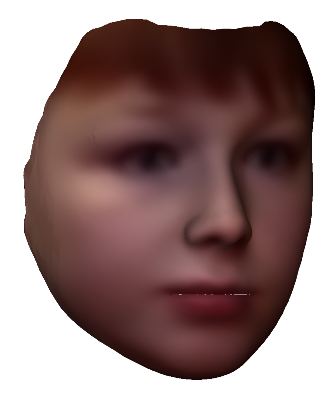} 
\includegraphics[width=0.08\linewidth,clip=true,trim=0 0 0 0]{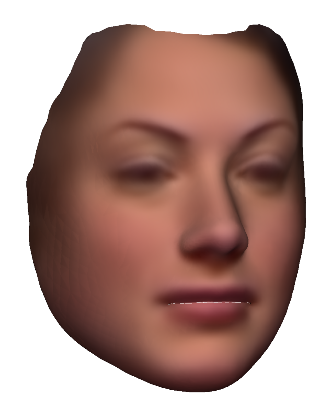} 
\includegraphics[width=0.08\linewidth,clip=true,trim=0 0 0 0]{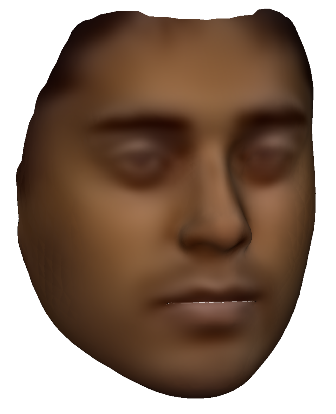} 

\caption{\emph{(Top row)}: Frame from the original video. \emph{(Second row)}: Reconstructed face texture using our real-time method. \emph{(Third row)}: Ground-truth face texture. \emph{(bottom row)}: Rendering from a novel pose.}
\label{fig:result_isomaps}
\end{figure*}

\subsection{Landmark Accuracy}

First, we evaluate the proposed approach on the ibug-Helen test set~\cite{Sagonas2016}, to be able to compare the landmark accuracy to other approaches in literature. We train a model using the algorithm of Section~\ref{sec:face_tracking}, using F-HOG features and 5 cascaded linear regressors in series.
On the official ibug-68 landmarks set, we achieve an average error of 0.049, measured in percent of the distance between the outer eye corners, as defined by the official ibug protocol (which they refer to as inter-eye-distance, \emph{IED}). The algorithm was initialised with bounding boxes given by the ibug face detector. Table~\ref{tbl:lmdet_results} shows a comparison with recent state of the art methods.

\begin{figure}[ht]
\begin{center}
    \includegraphics[width=1.0\linewidth]{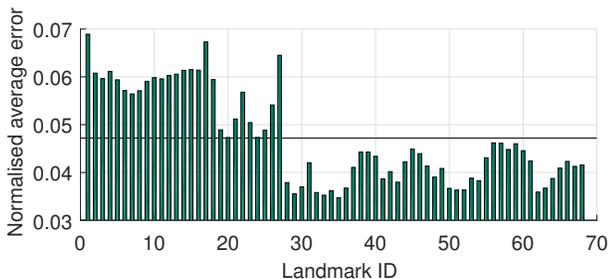}
\end{center}
\caption{Accuracy of each individual landmark on the 300-VW videos. 1-8 and 10-17 are contour landmarks, with significantly higher error. The horizontal bar depicts the average error.}
\label{fig:300vw_bar}
\end{figure}

To evaluate the accuracy of our tracking and the landmarks used for the shape reconstruction on in-the-wild videos, we evaluate the proposed approach on the public part of the 300-VW dataset~\cite{DBLP:conf/iccvw/ShenZCKTP15}. Across all videos, our tracking achieves an average error of 0.047.
Figure~\ref{fig:300vw_bar} shows the accuracy of each individual landmark.
Our approach achieves competitive results even on challenging video sequences.
Given that all 300-VW data is annotated semi-automatically, and the ground-truth contour landmarks are not well-defined and vary largely along the face contour, we believe this to be very close to the optimum achievable accuracy.

It is noteworthy that all the results in this paper were achieved by training on databases from different sources than 300-VW, no images from 300-VW were used in the training at any point.

\begin{table}[!t]
\renewcommand{\arraystretch}{1.3}
\caption{Landmark Error (in \% of IED)}
\label{tbl:lmdet_results}
\centering
\begin{tabular}{c||c|c|c}
\bfseries  & {\bfseries SDM}~\cite{xiong2013supervised} & {\bfseries ERT}~\cite{DBLP:conf/cvpr/KazemiS14} & \bfseries Ours\\ 
\hline\hline
HELEN & 0.059 & 0.049 & 0.049\\
300-VW & - & - & 0.047\\
\end{tabular}
\end{table}

\subsection{Face Reconstruction}

Our main experiment is concerned with reconstructing the 3D face and texture from in-the-wild video sequences. Since for such video sequences, no 3D ground-truth is available, we evaluate on the texture map, which account for shape as well as texture reconstruction accuracy.
We create a ground-truth isomap for ten 300-VW videos, by manually merging a left, frontal and right view, generated from accurate manual landmarks. We then compare our fully automatic reconstruction with these reference isomaps.

Figure~\ref{fig:result_isomaps} shows results of ibug 300-VW reconstructions. Our pipeline copes well with changing background, challenging poses, and, to some degree, varying illumination. The weighted fusion works well in these challenging conditions and results in a holistic, visually appealing reconstruction of the full face. Using an average-based fusion results in slight blur, but produces consistent results.

\subsection{Super-resolution Texture Fusion}

To evaluate the future potential of the proposed approach, we experiment with a median-based super-resolution approach to fuse the texture, assuming that we have knowledge of all frames of a video in advance.
We employ a simplified version of the technique proposed by Maier et al.~\cite{DBLP:conf/3dim/MaierSC15} for RGB-D data and adopt it to work in our scenario, resulting in a model-based super-resolution approach for texture fusion.
Instead of averaging as described in Section~\ref{sec:texture_reconstruction}, the fused colour value $ \hat{c} $ of a pixel is computed as the weighted median of all observed colour values with their associated weights $ O = \{ c_i, \omega_i \} $, and it is then computed as:

\begin{equation}
\hat{c} = \argmin_{c \in O} \sum_{\{ c_i, \omega_i \} \in O}\omega_i|c-c_i|
\label{eq:median_superres}
\end{equation}

\begin{figure}[ht]
\begin{center}
    \includegraphics[width=0.30\linewidth]{figures/big_fig_draft/002_isomap.png}
    \quad
    \includegraphics[width=0.30\linewidth]{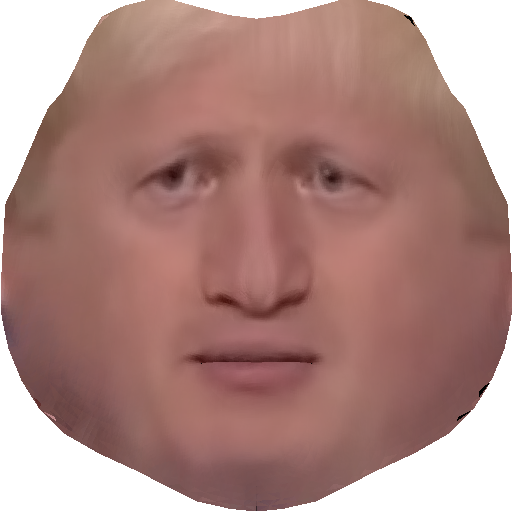}
\end{center}
\caption{\emph{(left)}: Average-based isomap reconstruction. \emph{(right)}: Reconstruction using median-based superresolution.}
\label{fig:superresolution}
\end{figure}

At the same time, while remapping the texture from the original frame to the isomap, we use a super-resolution scale factor of $ s = 2 $.
Figure~\ref{fig:superresolution} shows an example super-resolved isomap using this approach, computed offline.
The approach does not work in real-time yet and requires the whole video to be available in advance. We plan to extend the approach to work in an incremental manner on live-video streams.

\section{Conclusion} \label{sec:conclusion}
We presented an approach for real-time 3D face reconstruction from monocular in-the-wild videos.
The algorithm is competitive in landmark tracking and succeeds at reconstructing a shape and textural face representation, fusing different frames and view-angles.
In comparison with existing work, the proposed algorithm requires no subject-specific or manual training, reconstructs texture as well as a semi-dense shape, and it is evaluated on a true in-the-wild video database.

Furthermore, the 3D face model and the fitting library are available at 
\url{https://github.com/patrikhuber/eos}.
In future work, we plan to adopt the median-based super-resolution approach to work on real-time video streams.

\section*{Acknowledgments}
This work is in part supported by the Centre for Vision, Speech and Signal Processing of the University of Surrey, UK.
Partial support from the BEAT project (European Union's Seventh Framework Programme, grant agreement 284989) and the EPSRC Programme Grant EP/N007743/1 is gratefully acknowledged.

\IEEEtriggeratref{4}

\bibliographystyle{IEEEtran}
\bibliography{main}

\end{document}